%% file: main.tex
\documentclass{article}
\usepackage{spconf,amsmath,graphicx}
\usepackage{comment} 

\usepackage{microtype}
\usepackage{graphicx}
\usepackage{amsmath}
\usepackage{amssymb}
\usepackage{graphicx}
\usepackage{xcolor}
\usepackage[normalem]{ulem}
\usepackage{multirow}
\usepackage{booktabs}
\newcommand{\bs}[1]{\boldsymbol{#1}}
\usepackage{cite}

\newcommand{\ay}[1]{\textcolor{black}{#1}} 
\newcommand{\kw}[1]{\textcolor{black}{#1}} 
\newcommand{\claire}[1]{\textcolor{black}{#1}} 

\newcommand{\kaiwei}[1]{\textcolor{black}{#1}}
\newcommand{\thanh}[1]{\textcolor{black}{#1}} 

\title{Attentive Contextual Carryover for Multi-Turn \\End-to-End Spoken Language Understanding}
\name{\begin{tabular}{c}Kai Wei\textsuperscript{\rm *}, Thanh Tran\textsuperscript{\rm *}\thanks{\textsuperscript{\rm *} Equal contribution}, Feng-Ju Chang, Kanthashree Mysore Sathyendra, Thejaswi Muniyappa, \\
\textit{Jing Liu}, 
	\textit{Anirudh Raju}, 
	\textit{Ross McGowan}, 
	\textit{Nathan Susanj}, 
	\textit{Ariya Rastrow},
	\textit{Grant P. Strimel} \end{tabular}}

\address{Alexa Speech, Amazon}
\begin{document}

\maketitle

\begin{abstract}
\input{abstract.tex}
\end{abstract}
\section{Introduction}
\input{introduction.tex}

\vspace{-3pt}
\section{Problem Definition}
\vspace{-3pt}
\input{problem_definition.tex}

\label{context_s2i}

\vspace{-3pt}
\section{Proposed Contextual E2E SLU}
\input{context_encoder.tex}

\input{context_ingestion.tex}

\section{Experimental Settings}
\input{experiment.tex}

\section{Results}
\input{results.tex}

\section{Conclusion}
\input{conclusion.tex}



\bibliographystyle{IEEEbib}
\bibliography{refs}

\end{document}

%% file: abstract.tex
Recent years have seen \kw{significant}
advances in end-to-end \ay{(E2E)} spoken language understanding (SLU) \ay{systems}, \ay{which directly predict intents and slots from spoken audio.}
\ay{While dialogue history has been exploited to improve conventional text-based natural language understanding systems, current E2E SLU approaches have not yet incorporated such critical contextual signals in multi-turn and task-oriented dialogues.}
\ay{In this work, we propose a contextual E2E SLU model architecture that uses a \kw{multi-head} attention mechanism over encoded previous utterances and dialogue acts (actions taken by the voice assistant) of a multi-turn \kw{dialogue}. We detail alternative methods to integrate these contexts into the state-of-the-art recurrent and transformer-based models.}
\ay{When applied to a large de-identified dataset of utterances collected by a voice assistant, our method reduces \kw{average} word and semantic error rates by \kw{10.8\% and 12.6\%}, respectively. We also present results on a publicly available dataset and show that our method significantly improves performance over a non-contextual baseline.}

%% file: introduction.tex
{\kw{End-to-end (E2E) spoken language understanding (SLU) }
aims to infer intents and slots from \kaiwei{spoken audio} via a single neural network.
For example, when a user says \emph {order \kw{some} apples}, the model 
maps this spoken utterance (in the form of audio) to the intent \emph {Shopping} and slots such as \emph{\kw{Apple: Item}}.}
\kw{Recent research has made significant advances} in E2E SLU~\cite{lugosch2019speech,bhosale2019end,radfar2020end, lugosch2020using,  rongali2020exploring,rao2020speech}. 
Notably, \cite{rao2020speech}  \kaiwei{develops} a jointly trained E2E model, consisting of automatic speech recognition (ASR) and natural language understanding (NLU) models connected by a differentiable neural interface, \kaiwei{that outperforms the compositional SLU where ASR and NLU models are trained separately.}
Yet, how to incorporate contexts into E2E SLU remains unexplored.
 \begin{figure}[h!]
	\centering
	\includegraphics[height=6cm]{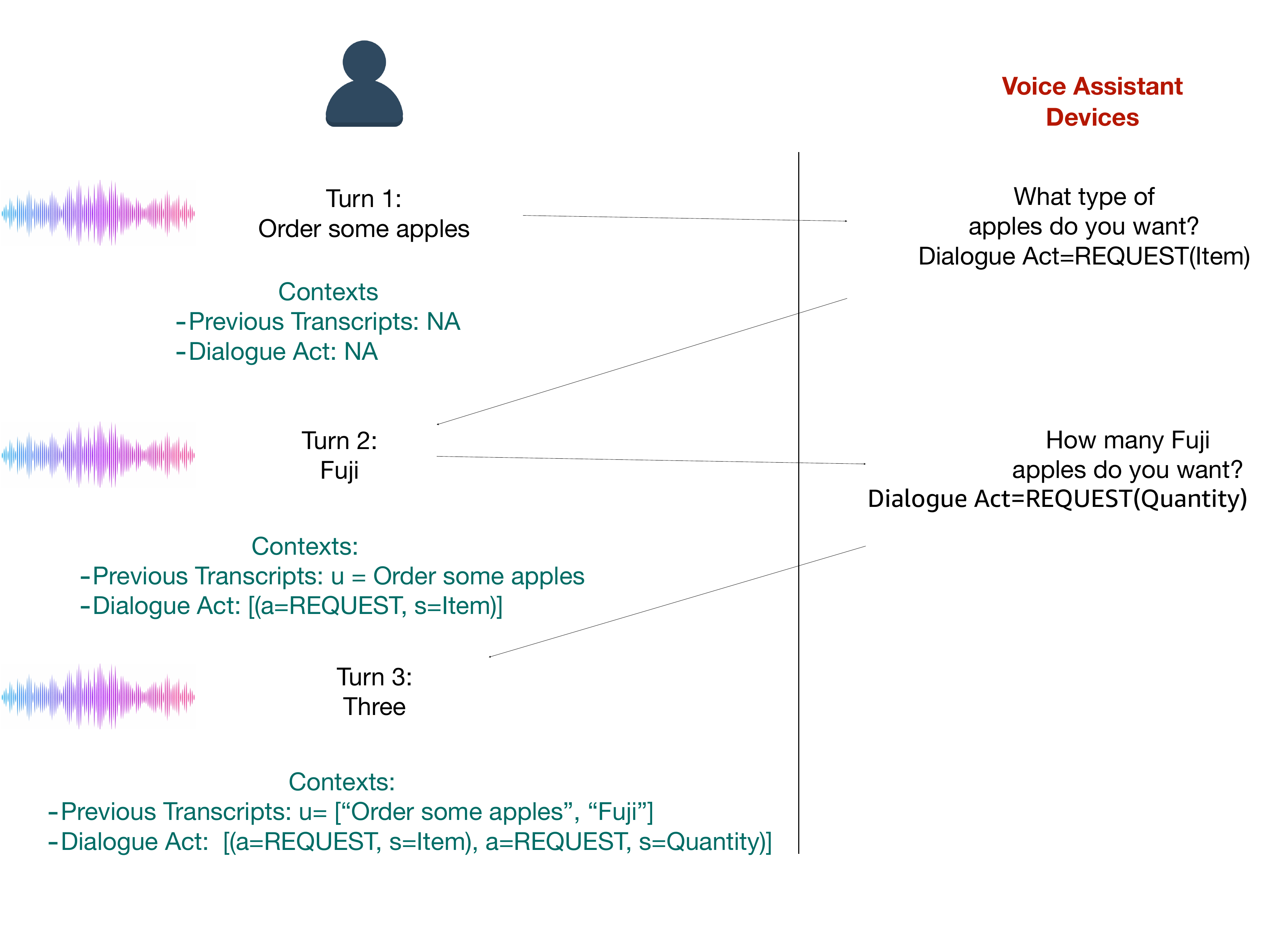}
        \vspace{-20pt}
	\caption{ {A} multi-turn dialog\kw{ue} example.}
	\label{fig:user-case}
	\vspace{-15pt}
\end{figure}

\kw{Contexts have been shown to significantly improve performance separately for ASR \cite{kim2018dialog, kim2019gated, kim2019cross, kim2020end, gupta2019casa, raju2018contextual,pundak2018deep,wu2020multistate,ray2021listen} and NLU \cite{gupta2019casa, Qin2021KnowingWT, su2019dynamically, abro2019multi, chen2019transfer, gupta2018efficient, chen2016end}.
For example, \cite{wu2020multistate}~\ay{\kaiwei{proposes} a multi-hot encoding to incorporate contextual information into \thanh{a} RNN transducer network (RNN-T) via the speech encoder sub-network and found that contexts such as dialogue state could improve accuracy of ASR.} \kaiwei{\cite{kim2019gated} uses cross-attention mechanism in an E2E speech recognizer decoder for two-party conversations.} 
\cite{gupta2018efficient}~has shown that encoding dialogue acts \ay{using} a feedforward network from dialogue history result\ay{ed} in a faster and more generalizable model without any \thanh{accuracy degradation} compared to \cite{chen2016end}. \cite{wang2019effective} encodes historical utterances with the BiLSTM and external knowledge with ConceptNet.} 



\kw{In this work, we propose a novel approach to encode dialogue history \thanh{in a} multi-turn E2E SLU \thanh{system}.}
  Figure \ref{fig:user-case}~\ay{illustrates a task-oriented turn-by-turn dialogue}
  between a user and a voice assistant (VA).
\kw{In this figure, the first turn is \emph{order some apples}. To \kaiwei{clarify the apple type, the VA asks \emph{What type of apples do you want?} at the second turn; and the user’s answer is \emph{Fuji}.} To \kaiwei{clarify the quantity,} the VA asks \emph{How many Fuji apples do you want?} at the third turn; and the user’s answer is \emph{three}. If \emph{three} is treated as a single-turn utterance, it is ambiguous since it \kaiwei{can} mean three apples or three o’clock. However, this utterance \kaiwei{can} correctly be interpreted as \emph{three apples} when presented with previous \kaiwei{dialogue contexts (e.g., \emph{order some apples} and \emph{Fuji}).}}
Prior E2E SLU research has focused on single-turn interactions where the VA receives the user's speech signals from just the current turn. They ignore~\kaiwei{the relevant contexts from previous turns that can enhance the VA's ability to correctly disambiguate user’s intent.}
 
 {\kw{In contrast to prior works, where dialogue acts are encoded singularly for ASR (e.g., \cite{wu2020multistate}) or NLU(e.g., \cite{gupta2018efficient}), we encode both dialogue acts and previous utterances to improve an E2E SLU architecture.}
\ay{Specifically, we propose a multi-head \kaiwei{gated} attention mechanism to encode dialogue context\kaiwei{s}. The attention-based context can be integrated at different layers of a neural E2E SLU model. We explore variants where either the audio frames, the neural interface layer (from ASR to NLU), or both are supplemented by the attention-based context vectors. Furthermore, \thanh{the learnable gating mechanism in our proposed multi-head gated attention can downscale the contribution of the context when needed}.}
}
\kw{O}ur proposed approach
 {\ay{improves} \kw{the performance of} \thanh{the} state-of-the-art E2E SLU models -- }\ay{namely} recurrent neural network transducer SLU and transformer transducer SLU
 on \ay{both internal} industrial voice assistant {datasets} and publicly available ones.

%% file: problem_definition.tex
\newcommand{\turn}{t}
\newcommand{\numturns}{T}
\newcommand{\dialogacts}{\mathcal{F}^\turn}
\newcommand{\dialogaction}{a}
\newcommand{\dialogslot}{s}
\newcommand{\dialogutterances}{\mathcal{U}^\turn}
\newcommand{\utterance}{u}
\newcommand{\actionset}{\mathcal{A}}
\newcommand{\slottypeset}{\mathcal{S}}
\newcommand{\numframes}{n}
\newcommand{\frames}{X^\turn}
\newcommand{\frameembedding}[1]{\bs{x^{t}_{#1}}}
\newcommand{\intentlabel}{y^\text{int.}}
\newcommand{\tokenlabel}[1]{y_{#1}^\text{tok.}}
\newcommand{\tokenlabels}{\{\tokenlabel{}\}}
\newcommand{\numtoks}{m}
\newcommand{\tokenlabelseq}{\{\tokenlabel{1}, \dots, \tokenlabel{\numtoks} \}}
\newcommand{\slotlabels}{\{y^\text{slot}\}}

We formulate the problem of \thanh{a} multi-turn E2E SLU as follows:  
In a multi-turn setting, a \textit{dialogue} between a user and the voice assistant system has $\numturns$ turns.
Each turn $\turn \in [1, \numturns]$ extends a growing list of dialogue acts $\dialogacts$=$\left\{(\dialogaction_1, \dialogslot_1),\dots, (\dialogaction_{t-1}, \dialogslot_{t-1})\right\}$ corresponding to the preceding system responses
and a list of the user's previous utterance transcripts $\dialogutterances$=$\{\utterance_1, \utterance_2, \dots, \utterance_{t-1}\}$. 
Each dialogue act $(\dialogaction_j, \dialogslot_j)$ in $\dialogacts$ \ay{comprises} of a dialogue action $\dialogaction_j$ from an action set $\actionset$  and a dialogue slot $\dialogslot_j$ from a slot set $\slottypeset$.
\ay{Take the second turn in Figure \ref{fig:user-case} as an example: the previous utterance $\utterance_{2} = \textit{Fuji}$, the dialogue action $\dialogaction_{2}={\textit{REQUEST}}$ and the dialogue slot $\dialogslot_{2} ={\textit{Item}}$}.
 
\noindent \textbf{Inputs and Outputs:} 
The inputs of each turn $\turn$ include acoustic input and dialogue contexts. The acoustic input $\frames$ comprises of
a sequence of $\numframes$ frame-based acoustic \ay{frames,}
 $\frames$=$\{\frameembedding{1}, \frameembedding{2}, \dots, \frameembedding{\numframes}\}$. Dialogue contexts include preceding dialogue acts $\dialogacts$,  \kw{and}
 the previous utterance transcripts $\dialogutterances$.
Our goal is to build a {contextual} neural E2E SLU architecture 
\ay{that correctly generates transcription and semantic outputs for each spoken turn, namely intent}
 $\intentlabel$,  transcript token sequence $\tokenlabels$, and slot sequence (one per token) $\slotlabels$.


%% file: context_encoder.tex
\newcommand{\actionpadlength}{l_a}
\newcommand{\actionembeddingmatrix}{M^\actionset}
\newcommand{\slotembeddingmatrix}{M^\slottypeset}
\newcommand{\embeddingdim}{d}
\newcommand{\dialogactionembedding}[1]{\bs{a_{#1}}}
\newcommand{\dialogslotembedding}[1]{\bs{s_{#1}}}
\newcommand{\fused}{g}
\newcommand{\fusedcaps}{G}
\newcommand{\fusedembedding}[1]{\bs{\fused_{#1}}}
\newcommand{\fusedembeddings}{G^{\turn}}
\newcommand{\fusedembeddingsaveraged}{\bs{\bar{\fused}^{\turn}}}
\newcommand{\linearfusionmatrix}{W^\fused}
\newcommand{\utterancepadlength}{l_b}
\newcommand{\utteranceembedding}[1]{\bs{u_{#1}}}
\newcommand{\utteranceembeddings}{U^{\turn}}
\newcommand{\utteranceembeddingsaveraged}{\bs{\bar{u}^{\turn}}}
\newcommand{\concatenatedembedding}{\bs{c^{\turn}}}
\newcommand{\concatrow}{\oplus}
\newcommand{\concatcol}{\oplus}
\newcommand{\contextembeddings}{C^\turn}
\newcommand{\attembedding}{\bs{c^\turn_{\text{att}}}}

\newcommand{\hiddens}{H^\turn}
\newcommand{\hiddenembedding}[1]{\bs{h^{t}_{#1}}}

\ay{The proposed} contextual E2E SLU architecture consists of \ay{a context encoder component, a }context combiner, and a base E2E SLU model.
The base model consists of ASR and neural NLU \ay{modules jointly trained via a differentiable neural interface~\cite{rao2021mean, rao2020speech}, which has been shown to achieve state-of-the-art SLU performance.}



\begin{figure}[h!]
	\centering
	\includegraphics[height=6.5cm, width=8cm]{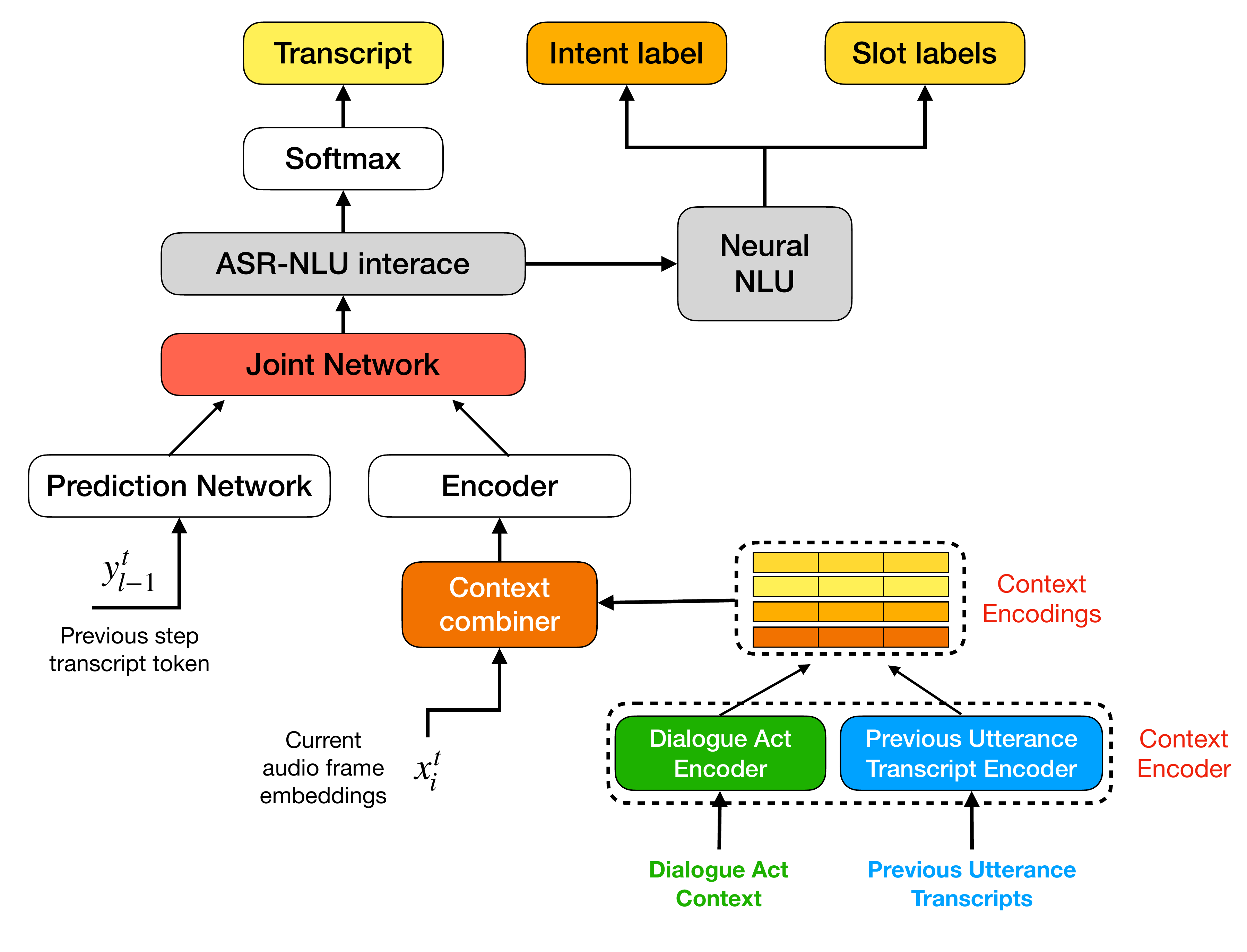}
	\vspace{-10pt}
	\caption{Contextual joint SLU model architecture using speech encoder context ingestion.}
	\label{fig:context-SLU}	
	\vspace{-6pt}
\end{figure}

Figure \ref{fig:context-SLU} shows the contextual E2E SLU model architecture using speech encoder context ingestion. \ay{The context encoder, described in Section~\ref{sec:context-encoder}, converts dialogue acts and utterance transcriptions of previous turns into contextual embeddings. The contextual embeddings are then combined with}
input audio features  $\frames$=$\{\frameembedding{1}, \frameembedding{2}, \dots, \frameembedding{\numframes}\}$ (described in Section~\ref{sec:context-combiner}) and then processed by the ASR module to {obtain the} output sequence $\bs{y}=\tokenlabelseq$, where the outputs $\tokenlabel{i}$ are \ay{transcription} graphemes, word or subword units \cite{graves2012sequence}.
\ay{Context} encoder embeddings are \ay{trained along with the rest of the E2E SLU architecture.}
The hidden interface (or ASR-NLU interface) \cite{raju2021end} is connected to the speech encoder via the joint network, which is a feedforward neural network that combines the outputs from the encoder and prediction network. 
This interface \ay{passes the intermediate hidden representation sequence} $\hiddens$=$\{\hiddenembedding{1}, \hiddenembedding{2}, \dots, \hiddenembedding{\numtoks}\}$ \ay{to a neural NLU module that} predicts intents $\intentlabel$ and a sequence of predicted slots, one per token, $\slotlabels$.
Our objective is to minimize the E2E SLU loss: ${\mathcal{L}_{\text{total}}} = {\lambda_1}{\mathcal{L}_{\text{tok.}}} +  {\lambda_2}{\mathcal{L}_{\text{slot}}}+{\lambda_3}{\mathcal{L}_{\text{int.}}}$, where $\mathcal{L}_{\text{tok.}}$ is the loss for word prediction, $\mathcal{L}_{\text{slot}}$ is the loss for slot prediction, and $\mathcal{L}_{\text{int.}}$ is the loss for intent prediction. The following section describes context encoder in detail.

\vspace{-3pt}
\subsection{Context Encoder}
\label{sec:context-encoder}
In this section, we describe approaches to encode dialogue acts and previous utterance transcripts. We first describe the \kaiwei{\textit{Dialogue Act Encoder}} that encodes the dialogue acts.
\kaiwei{Then, we describe the \textit{Previous Utterance Transcript Encoder} that} encodes transcripts from previous utterances.

\begin{figure}[h!]
	\centering
	\includegraphics[height=7cm, width=8.5cm]{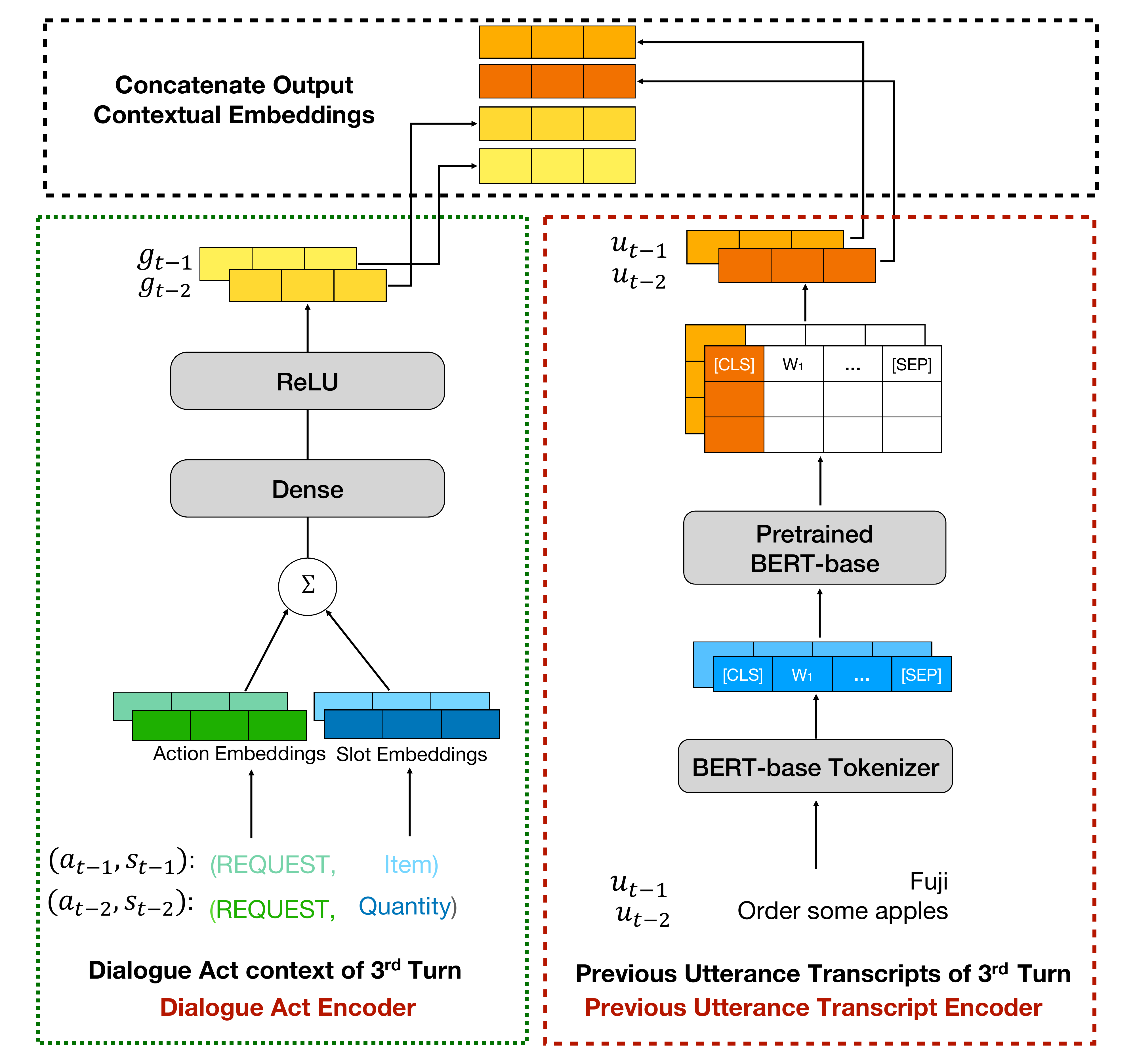}
	\vspace{-10pt}
	\caption{Encoding previous utterance transcripts and dialogue contexts in a multi-turn dialog between a user and a voice assistant system.}
	\label{fig:context-encoder}
	\vspace{-10pt}
\end{figure}

\vspace{-3pt}
\label{DialogAct}
\subsubsection{Dialogue Act Encoder}
\noindent\textbf{Input}: For the $\turn$-th turn, a list of dialogue acts for all previous turns denoted by
$\dialogacts$=$\left\{(\dialogaction_1, \dialogslot_1),\dots, (\dialogaction_{t-1}, \dialogslot_{t-1})\right\}$ is provided as \kaiwei{the} input.
We set {the} maximum number of dialogue action-slot pairs to $\actionpadlength$.
If $\dialogacts$ has less than $\actionpadlength$ dialogue action-slot pairs, we pad to length $\actionpadlength$ with a default action and slot. 

\noindent\textbf{Embedding layer}:
The embedding layer maintains two embedding matrices - a dialogue action embedding matrix $\actionembeddingmatrix \in \mathbb{R}^{\vert \actionset \vert \times \embeddingdim}$, 
and a dialogue slot embedding matrix $\slotembeddingmatrix \in \mathbb{R}^{\vert \slottypeset \vert \times \embeddingdim}$, 
with $\vert \actionset\vert$ and $\vert \slottypeset \vert$ referring to the total number of unique dialogue actions and slot types in the system, respectively. 
By passing each dialogue action $\dialogaction_j$ and dialogue slot $\dialogslot_j$ through their respective embedding matrices, we obtain their corresponding embeddings $\dialogactionembedding{j}$ and $\dialogslotembedding{j}$.

\noindent\textbf{Encoding layer}: {Given} the dialogue action and slot embeddings, $\dialogactionembedding{j}$ and $\dialogslotembedding{j}$, we fuse both embeddings via an element-wise addition followed by a nonlinear transformation with a $\mathit{ReLU}$ activation~\cite{gupta2018efficient} as summarized below.
\vspace{-3pt}
\begin{equation}
	\fusedembedding{j} = ReLU\big( \linearfusionmatrix (\dialogactionembedding{j}  + \dialogslotembedding{j}) \big)
	\label{equa:dialogAct-encoding}
\end{equation}

\noindent\textbf{Output}: 
{We produce the output $\fusedembeddings$ as a stack of dialogue act embeddings by aggregation of the list of $\fusedembedding{\turn - \actionpadlength}, \dots, \fusedembedding{\turn-1}$.


\label{PrevUtt}
\subsubsection{Previous Utterance Transcript Encoder}

\noindent\textbf{Input}: A list of previous utterance transcripts in the dialogue denoted by $\dialogutterances$=$\{\utterance_1, \utterance_2, \dots, \utterance_{\turn-1}\}$.
For each previous utterance transcript $\utterance_k$, we first tokenize it using the pre-trained BERT-base tokenizer.
Next, we prepend a [CLS] token and append a [SEP] token to the tokenized transcript.
We set the maximum number of previous utterance transcripts to $\utterancepadlength$. We pad empty sequences for $\dialogutterances$ if its length is less than $\utterancepadlength$, and take the $\utterancepadlength$ latest sequences in $\dialogutterances$ if its length is greater than $\utterancepadlength$.
\\
\noindent\textbf{Encoding layer:} From the tokenized transcripts, we apply the pre-trained BERT-\thanh{base} \cite{devlin2018bert} model to obtain an utterance transcript embedding $\utteranceembedding{k}$ for each previous utterance  $\utterance_k$ where we use the [CLS] token embedding as the summarized embedding for a full utterance transcript.
\\
\noindent\textbf{Output}: 
{Similar to $\fusedembeddings$, we output $\utteranceembeddings$ \ay{by stacking the list of utterance embeddings $\utteranceembedding{\turn-\utterancepadlength}, \dots, \utteranceembedding{\turn-1}$ from previous turns.}
 }


\begin{figure}[t]
	\centering
	\includegraphics[height=5.5cm, width=6.6cm]{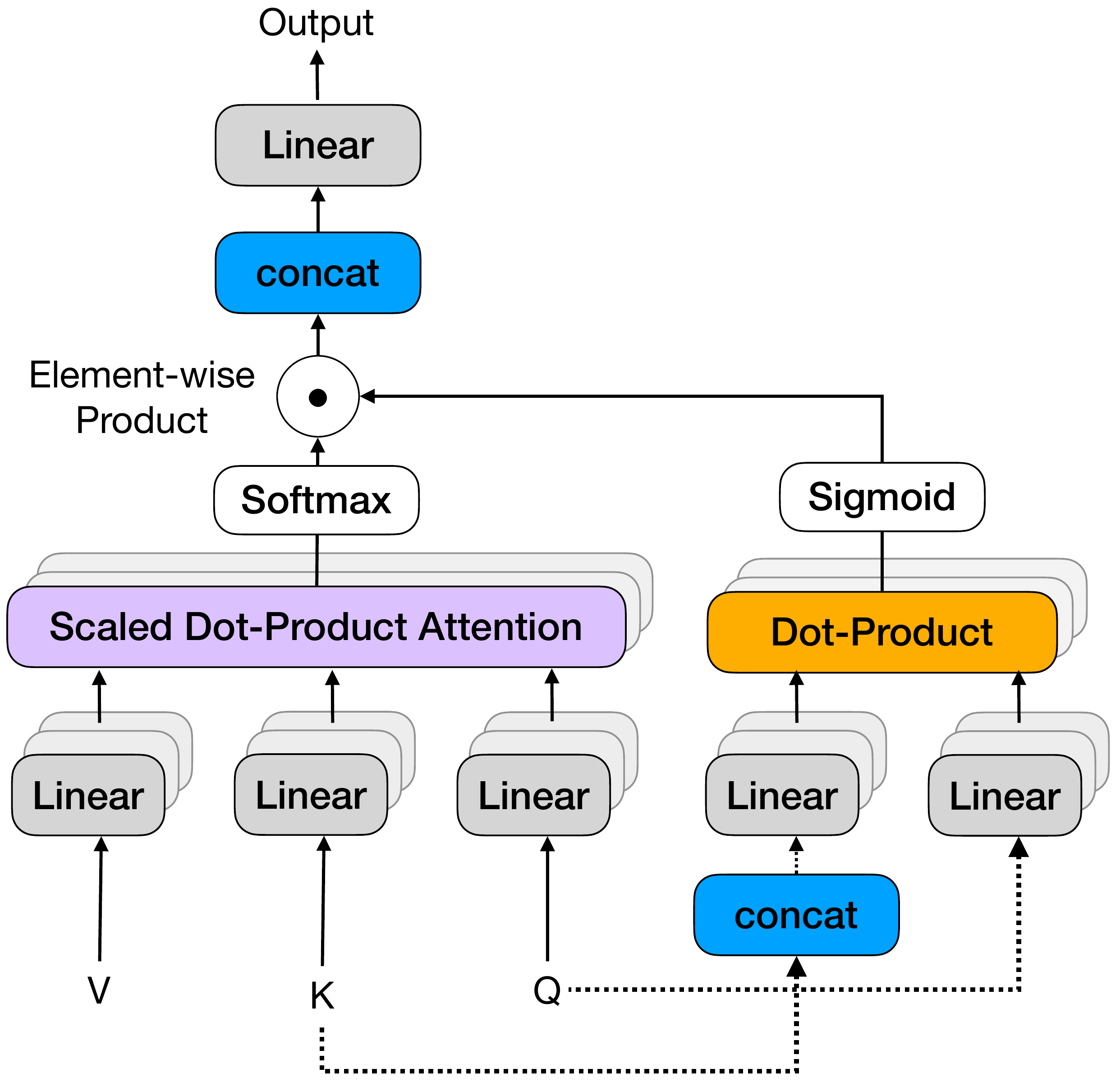}
	\vspace{-10pt}
	\caption{Architecture of Gated Multi-head attentions. 
}
	\label{fig:multigate-multihead-attention}
	\vspace{-15pt}
\end{figure}


\vspace{-5pt}
\subsection{Context Combiner}
\label{sec:context-combiner}
The context combiner combines the context encodings $\fusedembeddings$ and $\utteranceembeddings$ \ay{to create the final context vectors that are fed into the model.} 
We explore different ways to combine the context encodings into the model: (i) averaged contextual carryover, (ii) attentive contextual carryover, and (iii) gated attentive contextual carryover. 


To illustrate, we detail our approaches with an example that combines dialogue act encodings $\fusedembeddings$ and the previous utterance transcript encodings $\utteranceembeddings$ with the acoustic embeddings $\frames$=$\{\frameembedding{1}, \frameembedding{2}, \dots, \frameembedding{\numframes}\}$ of the $\turn$-th turn. 
Note that the same process can be applied to combine context encodings at different ingestion points in the model
(see Section \ref{context_ingest}).
We describe our context combiner methods below.
\vspace{-5pt}
\subsubsection{Averaged Contextual Carryover}
\label{sec:average-combiner}

Recall that $\fusedembeddings$ is the {stack} of dialogue act contextual embeddings and $\utteranceembeddings$ is previous utterance transcript embeddings at turn $t$. In this method, we first compute the average embeddings~\cite{wu2020multistate} of all dialogue act contextual embeddings $\fusedembedding{j} \in \fusedembeddings$ and average encodings of all previous utterance transcript embeddings $\utteranceembedding{k} \in \utteranceembeddings$. 
{Then, we combine the averaged contextual embeddings with the input by concatenating them with the acoustic embeddings, $\{\frameembedding{1}, \frameembedding{2}, \dots, \frameembedding{n}\}$, for each acoustic time step, \ay{as follows:}}

\vspace{-5pt}
\begin{equation}
\resizebox{0.7\linewidth}{!}{$
	\begin{aligned}
		\fusedembeddingsaveraged = \frac{1}{\actionpadlength} \sum_{\fusedembedding{j} \in \fusedembeddings} \fusedembedding{j} & \quad \utteranceembeddingsaveraged = \frac{1}{\utterancepadlength} \sum_{\utteranceembedding{k} \in \utteranceembeddings}  \utteranceembedding{k} \\
		\concatenatedembedding		   &= \left[ \fusedembeddingsaveraged ; \utteranceembeddingsaveraged \right] \\[0.75ex]
		\{\frameembedding{1}\prime, ..., \frameembedding{\numframes}\prime\} &= 
							\left\{
										\left[ \frameembedding{1} ; \concatenatedembedding \right] , \dots , \left[ \frameembedding{n} ; \concatenatedembedding   \right]
							\right\} 
	\end{aligned}
$}
	\label{equa:average-context-concat}
\end{equation}
\vspace{-5pt}

\vspace{-5pt}
\subsubsection{Attentive Contextual Carryover}
\label{sec:gated-attention}
\ay{Averaging contextual embeddings of the previous turns can hamper the ability of the model to access fine-grained contextual information for a specific turn and time step.}
\ay{Therefore,} we \thanh{utilize} the multi-head attention \ay{mechanism \cite{vaswani2017attention}, which uses acoustic embeddings, of each time step, to attend to relevant dialogue contexts and create the final contextual embeddings. }

\claire{Specifically, we create the queries, keys, and values, $Q_i, K_i, V_i, i \in \{g,u\}$ via \ay{linear projections}
as follows:}
\claire{\vspace{-5pt} 
\begin{align}
Q_g = W_{g}^{(q)} X^t \text{;} \;\; K_g = W_{g}^{(k)} G^t \text{;} \;\; V_g = W_g^{(v)} \fusedembeddings \nonumber \\
Q_u = W_u^{(q)}  X^t \text{;} \;\; K_u = W_u^{(k)}  U^t \text{;} \;\; V_u = W_u^{(v)}  \utteranceembeddings
\label{equa:query-key-value}
\end{align}}
Here, $X^t$, $\fusedembeddings$, $ \utteranceembeddings$ are acoustic, dialogue act, and previous utterance embeddings for the $t$-th turn, respectively. Matrices $W_{g}^{(\cdot)}, W_{u}^{(\cdot)}$ \ay{are learnt linear projections.}
}
\claire{
\ay{A scaled dot-product attention is then used to calculate the final dialogue and utterance context vectors through the weighted sum of projected contextual embeddings of the previous turns.} This process is formulated as:} \claire{\vspace{-5pt}
\begin{align}
\bs{\alpha_g} = \text{Softmax} \bigg(  \frac{Q_g K_g^\top}{\sqrt{d}}\bigg), \quad C^t_{g} = \bs{\alpha_g} V_g \nonumber \\
\bs{\alpha_u} = \text{Softmax} \bigg(  \frac{Q_u K_u^\top}{\sqrt{d}}\bigg), \quad C^t_{u} = \bs{\alpha_u} V_u
\label{equa:scale-dot-attention}
\end{align} }
where $d$ is the hidden size of the attention layer applied for numerical stability \cite{vaswani2017attention}. The attention outputs $C^t_{g}$ and $C^t_{u}$  are then concatenated with the acoustic embeddings $\frames$ provided as input.

\subsubsection{Gated Attentive Contextual Carryover}
\label{sec:gated-attention2}
One limitation of the attention mechanism is that it cannot downscale the contribution of a context when needed~\cite{xue2020not}.
Take a two-turn dialogue as an example: 

\vspace{1mm}
\hangindent=0.5cm{A user asks a voice assistant to \emph{call uncle sam} in the first turn, and the system confirms back to see if the user wants to call \emph{Uncle Sam's Sandwich Bar} (associated dialogue act is REQUEST(restaurant)). Then, in the second turn, the user corrects that she wants to ``call my uncle sam''.}
\vspace{1mm}

\noindent
In this case, simply applying multi-head attention as described in Eq.(\ref{equa:query-key-value})-(\ref{equa:scale-dot-attention}) on the previous turn utterance \textit{call uncle sam}, $U^t$, and dialogue act \textit{REQUEST(restaurant)}, $G^t$, can lead to \thanh{a} wrong interpretation \thanh{for} the second turn. This is because the results of the Softmax function in Eq.(\ref{equa:scale-dot-attention}) assigns dialogue act context to positive scores, misleadingly associating \emph{uncle sam} with a \emph{restaurant name} rather than a \emph{person name}.
 
\thanh{Inspired by the gating mechanism  to control information flow or integrate different types of information \cite{kim2019end,chung2014empirical,hochreiter1997long,kim2018towards,kiros2018illustrative,kim2019gated}}, we introduce a learnable gating mechanism on top of the \thanh{multi-head} attentive contextual carryover to further reduce a context's influence when it does not help the interpretation.  
Specifically,  we concatenate all the contextual embeddings in $\fusedembeddings$ and $\utteranceembeddings$ to obtain $C^t_{c}$.
Then, we \claire{obtain} the gating scores \claire{by computing} the similarity between the linearly projected $\frames$ and $C^t_{c}$, as follows:
\vspace{-5pt}
\begin{equation}
	\beta_{c} = \text{sigmoid}\bigg( Q_c K_c^\top \bigg),
	\label{equa:gating}
\end{equation}
\begin{equation}
	\nonumber
	Q_c= W_c^{(q)}X^t \text{;} \;\; K_c = W_c^{(k)} {C^t_c}
	\label{equa:gated-query-key}
\end{equation}
where $W_c^{(q)}$ and $ W_c^{(k)}$ are learnable parameters. 
\claire{$\beta_{c} \in  \mathbb{R}^{n \times 1}$ and $n$ is the number of frames.}
Each entry $\beta_{c}$ 
shows how much 
contexts contribute to the acoustic embedding $\frameembedding{i}$ at $i$-th frame, $i \in \left[\thanh{1,}\; n \right]$). 
We replicate $\beta_{c}$ to make it have the same dimension as $\bs{\alpha}_g$ and $\bs{\alpha}_u$. The gated attention scores $\bs{\gamma}$ are then computed by the element-wise product between $\bs{\alpha}$ scores  and $\bs{\beta_{c}}$:
\begin{equation}
\begin{aligned}
	\bs{\gamma_g} = \bs{\alpha_g} \odot \bs{\beta_{c}} \quad \bs{\gamma_u} = \bs{\alpha_u} \odot \bs{\beta_{c}}
	\label{equa:gated-attentions}
\end{aligned}
\end{equation}

We compute the gated attentive contextual embeddings across each attention head, as follows: 

\begin{equation}
	\begin{aligned}
		C^t_{g, \text{gated}} =  \bs{\gamma_g} {V_g} {;} \;\; C^t_{u,\text{gated}} = \bs{\gamma_u} V_u.
	\end{aligned}
\end{equation}

Finally, $C^t_{g, \text{gated}}$ and $C^t_{u,\text{gated}}$ are row-wise concatenated with the acoustic embeddings $\frames$ as input.

%% file: context_ingestion.tex
\vspace{-5pt}
\label{context_ingest}
\subsection{Context Ingestion Scenarios}

We consider the integration of the context encoder using three schemes: ingestion by the speech encoder network,   ingestion with the hidden ASR-NLU interface, and finally at both insertion points.

\noindent\textbf{Speech encoder ingestion:} In this method, we incorporate the outputted context embeddings only into the acoustic embeddings for ASR pre-training/training task.
This approach is motivated by prior research showing that context benefits the speech encoder more than the prediction network of ASR transducer models ~\cite{ray2021listen} .
To combine context with acoustic embeddings, we input the acoustic embeddings $\frames$=$\{\frameembedding{1}, \frameembedding{2}, \dots, \frameembedding{\numframes}\}$ as the query, and the context encodings $G^t$, $U^t$ serve as the keys and values in the context combiner.
The output $\{\bs{x^{t}_{1}}\prime, ..., \bs{x^{t}_{n}}\prime\}$ with ingested context (Equation ~(\ref{equa:average-context-concat})) are then used to perform the ASR task.

\noindent\textbf{ASR-NLU interface ingestion:}  In this approach, we ingest the output context embeddings only into the ASR-NLU interface embeddings for the SLU training task.
As such, we now use the ASR-NLU interface embeddings $\hiddens$=$\{\hiddenembedding{1}, \hiddenembedding{2}, \dots, \hiddenembedding{\numtoks}\}$ as queries for context combiner instead of the acoustics.

\noindent\textbf{Shared context ingestion:} In this method, we integrate context into both acoustic embeddings and ASR-NLU interface embeddings.
We maintain a shared context encoder between the ASR and NLU submodule, resulting in a shared $G^t$, $U^t$ between them.
For fusion, we maintain two separated context combiners to increase the context ingestion flexibility.
Specifically, we establish a gated multi-head attentive context combiner for the ASR submodule with $\frames$ as queries, while having another gated multi-head attentive context combiner for the NLU submodule with $\hiddens$ as queries.


In the following sections, we perform experiments on incorporating multi-turn context into two SLU architectures: a Transformer-based Joint SLU model and an RNN-T based Joint SLU model.

%% file: experiment.tex
\subsection{Datasets}
The internal industrial voice assistant (IVA) dataset is a far-field dataset with \kw{more than 10k hours of audio data} and their corresponding \kw{intent and slot annotations.}
It is a multi-domain dataset with both single-turn and multi-turn utterances.
In total, there are 55 intents, 183 slot types, and 49 dialogue acts.
In addition, we \kw{built} a synthetic and publicly available multi-turn E2E SLU (Syn-Multi) dataset based on~\cite{shah2018building}.
\cite{shah2018building}~contains two datasets with a text-only format from Restaurant (11,234 turns in 1,116 training dialogues) and Movie (3,562 turns in 384 training dialogues) domains. To obtain audio signals, we used a Transformer text-to-speech model~\footnote{https://github.com/as-ideas/TransformerTTS} to synthesize the audio and combine the two datasets into one dataset for model training and evaluation. Finally, we \kw{used} SpecAugment~\cite{park2019specaugment} to augment audio feature inputs. In total, Syn-Multi has 3 intents, 12 slot types, and 21 user dialogue act types\footnote{https://github.com/google-research-datasets/simulated-dialogue}. 

\subsection{Implementation setup}

\noindent\textbf{Audio features:} The input audio features 
are
64-dimensional LFBE features extracted every 10 ms with a window size of 25 ms from audio samples. The features of each audio frame are 
stacked with the features of two previous audio frames, followed by a downsampling factor of 3 to achieve a low frame rate, resulting in 192 feature dimensions per audio frame. 
We use a token set 
\kw{with}
4,000 wordpieces trained \kaiwei{by} \thanh{the} sentence-piece tokenization model~\cite{kudo2018subword}. 

\noindent\textbf{Model setup:} Table \ref{tbl:stats} shows \thanh{our} model setup details. \kw{We built contextual E2E SLU models based on the Recurrent Neural Network Transducer (RNN-T)~\cite{he2019streaming}  and the Transformer Transducer (T-T)~\cite{zhang2020transformer}, respectively.
\kw{E2E SLU models} share
an audio encoder network \kw{that} 
encodes 
LFBE features, a prediction network \kw{that} 
 encodes a sequence of predicted wordpieces,
a joint network that combines \kw{the encoder and the prediction network,}
and an NLU tagger \kw{that predicts intents and slots.} The intent tagger contains two feedforward layers before projecting into the number of intents, and the slot tagger directly takes the output embeddings from the NLU tagger and projects them into the slot size. The audio encoder in the E2E T-T SLU and E2E RNN-T SLU are Transformer layers (with 4 attention heads) and LSTM layers, respectively. The NLU tagger in E2E T-T SLU and E2E RNN-T SLU are transformer layers (with 8 attention heads) and BiLSTM layers, respectively. For $\actionpadlength$ (the maximum number of dialog action-slot pairs) and  \thanh{$\utterancepadlength$ (the maximum number of previous utterance transcripts)},  we set \kaiwei{$\actionpadlength$ = $\utterancepadlength$ = 5} in the IVA dataset.
We set \kaiwei{$\actionpadlength$ = $\utterancepadlength$ = 20} in the {Syn-Multi} dataset.} 


\noindent\textbf{Training setup:}
We adopt a stage-wise \kw{joint} training strategy for \kw{the proposed contextual models and baseline non-contextual models.}
We first pre-trained an ASR model to minimize \thanh{the} RNN-T loss~\cite{graves2012sequence}. \kw{We then freeze \thanh{the} ASR \thanh{module} to train the NLU module to minimize the cross entropy losses for the intent and slot predictions.~\kaiwei{During training, all constituent subwords of a word are tagged with its slot. During inference, the constituent subwords are combined to form the word, and the slot tag for the last constituent subwords is taken as the slot tag for the word.}} Last, \kw{we jointly tuned ASR and NLU \thanh{modules} to \kaiwei{minimize} all three losses.}~\kaiwei{We used the teacher forcing technique~\cite{williams1989a} that uses the human-annotated transcripts of previous turns for training, and the automatic transcripts of previous turns from our model for inference.}
We applied the Adam optimizer~\cite{kingma2014adam} for all model training. For E2E RNN-T SLU, the learning rate is warmed linearly from 0 to $5 \times 10^{-4}$ during the first 3K steps, held constant until 150K steps, and then decays exponentially to $10^{-5}$ until 620K steps. For E2E T-T SLU, the learning rate is warmed from 0 to $5 \times 10^{-4}$ in the first 16K steps, then is decayed to $10^{-5}$ in the following 604K steps exponentially.
We used 24 NVIDIA® V100 Tensor Core GPUs and a batch size of 32 for training \thanh{the} model. 
\vspace{-5pt}
\begin{table}[h!]
\resizebox{\linewidth}{!}{
	\begin{tabular}{lcc|ccl}
		\hline
		\multicolumn{1}{c}{}   & \multicolumn{2}{c|}{IVA Dataset} & \multicolumn{2}{c}{Syn-Multi Dataset} \\
		Statistic              & \multicolumn{1}{c}{RNN-T SLU} & \multicolumn{1}{c|}{T-T SLU} & \multicolumn{1}{c}{RNN-T SLU} & \multicolumn{1}{c}{T-T SLU} \\
		\hline
		\textbf{Audio encoder network}      &  				&     		&         &           		 \\
		\;\;\;  \# Layers								 & 5         &  	   6   &  4	    &    2      		  \\
		\;\;\;  Layer embed-size				   & 736     &  256       & 640  &    256     	   \\
		\;\;\;  \# Attention heads					 & -- 	    &     4        & --     &    4      		  \\
		\;\;\;  \#FeedForward layer     		  & 1          &     1       & 1 		&    1       		 \\
		\;\;\;  FeedForward embed-size 	      & 512    &  2048     & 256 	&   512     		    \\
		\hline
		\textbf{Prediction network}             &  			&     	       &  			&            			\\
		\;\;\; \# Layers								      & 2    	&  	   2  	  &  2   	&   2           \\
		\;\;\;  Layer embed-size				    & 736     &  	736 &   640  &   640           \\
		\;\;\;  \#FeedForward layer     		   & 1         &   1    	& 1 		&    1       		 \\
		\;\;\;  FeedForward embed-size 	      & 512     &   512    &  256	&    512    		    \\
		\hline
		\textbf{Joint network}          		   &  			 &     	     &  		&            			\\
		\;\;\;  Vocab embed-size				  & 512     &     512   & 512  & 512                 \\
		\;\;\;  \#FeedForward layer     		   & 1         &   1        & 1 		& 1          		 \\
		\;\;\;  FeedForward embed-size 	      & 512     &   512    & 512 	&   512     		    \\
		\;\;\;  Activation								 & tanh   & tanh     &  tanh  &	tanh			\\
		\hline
		\textbf{NLU decoder network}	    &           &      	      &  	   	 &            				\\
		\;\;\;  \# Layers           			  		   & 2         &    2     &  2	     &  2          			\\
		\;\;\;  Layer embed-size        		   	& 256     &   256  &  256   & 256            	\\
		\;\;\;  \#FeedForward layer     		   & 1         &   1        &  1	  &  1         		 \\
		\;\;\;  Feedforward size       			    & 256      &   256   & 256	 &     256       		\\
		\;\;\; \textbf{Intent Predictor Network}        &          &       	 	&   	   &           		 \\
		\;\;\;\;\;\;	\;\;\;  \#FeedForward layer & 2        &   2      &  2		&   2        		 \\
		\;\;\;\;\;\; \;\;\;  Feedforward size         & 512    &   512   &  512	 &  512          		\\
		\;\;\;\;\;\; \;\;\;  Activation                  & relu    &   relu    & relu    &   relu        		\\
		 \;\;\;\;\;\;	\;\;\;  \#FeedForward layer & 1        &   1      &  1		&   1        		 \\
		\;\;\;\;\;\; \;\;\;  Feedforward size         & \#intent    &   \#intent   &  \#intent	 &  \#intent          		\\
		\;\;\; \textbf{Slot Tagger Network}        &          &       	 	&   	   &           		 \\
		\;\;\;\;\;\;	\;\;\;  \#FeedForward layer & 1        &   1      &  1		&   1        		 \\
		\;\;\;\;\;\; \;\;\;  Feedforward size         & \#slots    &   \#slots   &  \#slots	 &  \#slots          		\\
		\;\;\; \;\;\; \;\; \# Attention heads					 & -- 	    &     8      & -- 	  &  8        		  \\
		\hline
	\end{tabular}
}
	\caption{Model setup for E2E SLU.} 
	\label{tbl:stats}
	
\end{table}
\vspace{-17pt}

\subsection{Evaluation Metrics and Baselines}
We evaluate the model performance on word error rate (WER), intent classification error rate (ICER), and semantic error rate (SemER). \kw{WER measures the proportion of words that are misrecognized (deleted, inserted, or substituted) in the hypothesis relative to the reference. ICER measures the proportion of utterances with a misclassified intent. SemER combines intent and slot accuracy into a single metric, i.e., SemER = \# (slot errors + intent errors) / \# (slots + intents in reference).}
We only show relative error rate reduction results on the IVA dataset. 
Take WER for example, given a method A’s WER ($\text{WER}_A$) and a baseline B’s WER ($\text{WER}_B$), the relative word error rate reduction (WERR) of A over B can be computed by $(\text{WER}_B - \text{WER}_A) / \text{WER}_B$; the higher the value, the greater the improvement. 
\kw{We denote relative errors for WER, ICER and SemER as WERR, ICERR and SemERR.}


%% file: results.tex

\textbf{Improving E2E SLU with contexts\claire{:}} Table~\ref{tbl:overall} shows overall model performance and the total number of parameters (in millions) of the baseline and our proposed models on the IVA dataset. We observe that contexts play a crucial role in improving E2E SLU across speech recognition and semantic interpretations.
Particularly, our contextual E2E RNN-T SLU model relatively reduces 7.75\% of WER, 10.96\% of ICER, and 14.56\% of SemER. Our contextual E2E T-T SLU model relatively reduces 13.83\% of WER, 11.06\% of ICER, and 10.60\% of SemER.
Interestingly, encoding contexts with gated attentive contextual carryover performed better than the traditional multi-head attention \cite{vaswani2017attention}. It gave the best performance with a relative improvement for SemER of 14.56\% and 10.6\% respectively across RNN-T and T-T based models. 

For all subsequent discussion, we focus on SemER, as it summarizes the performance across all tasks.

\vspace{-5pt}
\begin{table}[h!]
	\resizebox{\columnwidth}{!}{%
		\begin{tabular}{rrrrr}
			\multicolumn{2}{r}{} &  \multicolumn{3}{c}{Relative Error Reduction} \\ 
			\cline{3-5}
			Model & Config. (\# params) & {WERR} & {ICERR} & {SemERR}  \\ \hline
			& No Context (35.12M)  & Baseline       & Baseline          & Baseline            \\ 
			& w/ DA (35.31M) & 5.86\%         & 8.94\%  & 8.06\%      \\
			E2E  
			& PrevUtt (37.38M)& 7.44\%       & 6.62\%  & 12.23\%            \\ 
			 RNN-T& DA+ PrevUtt + AvC (37.57M) & 7.38\%      & 8.74\%  & 12.66\%          \\ 
			 SLU & DA + PrevUtt + AttC  (37.72M) & 7.88\%         & 6.92\%          & 13.14\%            \\ 
			& DA + PrevUtt + GAttC (37.94M) & 7.75\%       & 10.96\%  & 14.56\%          \\  \hline
			& No Context (28.58M) & Baseline         & Baseline          & Baseline            \\ 
			&w/ DA (28.61M) & 5.39\%       & 4.59\%  & 1.48\%       \\  
			E2E         
			& PrevUtt (28.78M) & 12.37\%          & 8.87\%  & 6.32\%           \\ 
			 T-T& DA+ PrevUtt + AvC (28.80M) & 11.50\%      & 8.46\%  & 7.85\%            \\ 
			 SLU&  DA + PrevUtt + AttC (30.67M)   & 12.63\%         & 9.50\%          & 9.27\%       \\ 
			& DA + PrevUtt + GAttC (30.89M) & 13.83\%       & 11.06\%  & 10.60\%           \\  \hline
		\end{tabular}
	}
	\vspace{-10pt}
	\caption{Overall results on the IVA dataset. NoContext: E2E without contexts. DA and PrevUtt: dialogue act and previous utterance context. AvC: average contextual carryover.  AttC: attentive contextual carryover.  GAttC: AttC with gating layers.}
	\label{tbl:overall}
\end{table}



Table~\ref{tbl:mutl} summarizes the results for utterances with two turns, three turns, and at least four turns. We observe that encoding contexts can lead to an average relative improvement of 40.73\% and 37.09\% across RNN-T and T-T E2E SLU.
 \vspace{-5pt}
\begin{table}[h!]
 \centering
    \resizebox{\columnwidth}{!}{%
    \begin{tabular}{rrrrr}
      \multicolumn{2}{c}{{}} & \multicolumn{3}{c}{{SemERR}} \\  
      \cline{3-5}
     Model & Config. & 2-turn & 3-turn & 4-turn + \\ \hline
   			   E2E RNN-T  SLU                            & No Context   & Baseline         & Baseline           & Baseline        \\ 
			          & DA + PrevUtt + GAttC  &     30.35\%     &  37.80\% & 54.04\%           \\ \hline
			  E2E T-T 	SLU		       & NoContext   & Baseline         & Baseline          & Baseline      \\ 
			                   & DA + PrevUtt + GAttC &      37.07\%    &  37.91\%&  36.30\%    \\ \hline
    \end{tabular}
    }
    \caption{Results on the IVA multi-turn utterances.}
    \label{tbl:mutl}
\end{table}


%

\noindent{\textbf{The effect of context ingestion:}} \kaiwei{Table \ref{tbl:ingestion} and Table \ref{tbl:sdm} show the effects of context ingestion on the E2E SLU performance.}
We observe that the context encoder improves E2E SLU for all scenarios, giving an average relative improvement of 13.69\% and 12.07\%, respectively, across RNN-T and T-T E2E SLU.
Compared to the speech encoder and hidden interface ingestion, \thanh{the} shared context ingestion gave the biggest improvement on T-T E2E SLU with a relative improvement of 19.4\%. 
\begin{table}[h!]
 \centering
    \resizebox{\columnwidth}{!}{%
    \begin{tabular}{rrrrr}
      \multicolumn{5}{r}{Relative Error Reduction} \\  
      \cline{3-5}
     Model & Config & WERR & {ICERR} & {SemERR}  \\ \hline
   			                              			& No Context   & Baseline           & Baseline            & Baseline            \\ 
			{E2E }                & Speech Encoder & 7.75\%         & 10.96\%  & 14.56\%            \\ 
			RNN-T		& ASR-NLU Interface & 8.83\%         & 8.23\%  & 13.56\%            \\ 
                     	SLU		 	& Shared Context & 9.14\%         & 7.13\%  & 12.95\%      \\ \hline
                          						& No Context   & Baseline           & Baseline            & Baseline            \\ 

			{E2E}            & Speech Encoder & 13.83\%         & 11.06\% & 10.6\%      \\ 
			 T-T		    & ASR-NLU Interface  & 1.26\%         & 6.89\%  & 6.21\%       \\ 
			 SLU		    & Shared Context & 15.16\%         & 20.81\% & 19.4\%      \\  \hline

    \end{tabular}
    }
	\vspace{-10pt}
    \caption{The effect of context ingestion: IVA datasets.}
    \label{tbl:ingestion}
\end{table}
\vspace{-10pt}
\begin{table}[h!]
	\centering
	\resizebox{\columnwidth}{!}{%
		\begin{tabular}{rrrrr}
			\multicolumn{5}{r}{Absolute Error Rate} \\ 
			\cline{3-5}
			Model & Config & WER & ICER & SemER  \\ \hline
			& No Context   			    	& 16.02\%        & 32.49\%   &  40.76\%           \\ 
			{E2E}   	        & Speech Encoder  	& 19.76\%        & 3.62\% & 29.24\%            \\ 
			RNN-T & ASR-NLU Interface  	& 10.59\%        & 0.36\% & 18.99\%             \\ 
			 SLU & Shared Context	& 12.14\%       & 0.25\%  & 18.55\%      \\ \hline
			& No Context   				& 13.06\%      & 30.4\%  &          36.83\%          \\ 
			{E2E}                        & Speech Encoder 	& 14.1\%         & 2.38\%  & 26.56\%       \\ 
			 T-T & ASR-NLU Interface  	& 12.81\%       & 0.25\%  & 18.49\%       \\ 
			SLU & Shared Context 	& 13.68\%      & 0.21\%  &18.62\%      \\  \hline
			
		\end{tabular}
	}
	\vspace{-10pt}
	\caption{The effect of context ingestion: {Syn-Multi} datasets.}
	\label{tbl:sdm}
\end{table}

\kw{We also qualitatively examined the effect of contexts.}
Contextual models recognized \emph{cancel} correctly with the \emph{Select(Time)} dialogue act context, whereas non-context model recognized the word as \emph{cascal}. 
Further, contextual models can better handle ambiguous utterances. For example, contextual models correctly predict utterance \emph{next Monday for inferno} as \emph{BuyMovieTickets} intent \kw{as} its previous utterance is \emph{i want to buy movie tickets}, whereas non-context models confuse this utterance with \emph{ReserveRestaurant} intent.

%% file: conclusion.tex
We propose a novel E2E SLU approach \kaiwei{where a multi-head gated attention mechanism is introduced to \thanh{effectively incorporate the} dialogue history from \thanh{the} spoken audio.}  Our proposed approach significantly improves E2E SLU accuracy on the internal industrial voice assistant and publicly available datasets compared to the non-contextual E2E SLU models. In the future, we will apply our proposed approach on other datasets and further improve our contextual model architecture.